\documentclass[11pt,a4paper]{article}
\usepackage[hyperref]{conll-2019}
\usepackage{hyperref}
\usepackage{times}
\usepackage{latexsym}
\usepackage{CJKutf8}
\usepackage{caption}
\usepackage{subcaption}
\usepackage{graphicx}
\usepackage{url}
\usepackage{booktabs}
\usepackage{float}
\usepackage{amsmath}
\usepackage{multirow}
\newcommand\norm[1]{\left\lVert#1\right\rVert}
\usepackage{bm}

\aclfinalcopy % Uncomment this line for the final submission

\newcommand{\eg}{{e.g.}}
\newcommand{\ie}{{i.e.}}

\newcommand{\hhide}[1]{}
\newcommand{\eat}[1]{\ignorespaces}

\newcommand\blfootnote[1]{%
  \begingroup
  \renewcommand\thefootnote{}\footnote{#1}%
  \addtocounter{footnote}{-1}%
  \endgroup
}

\title{Improving Pre-Trained Multilingual Models with Vocabulary Expansion}

\author{
 Hai Wang\textsuperscript{1}\textsuperscript{*} ~~ Dian Yu\textsuperscript{2} ~~ Kai Sun\textsuperscript{3}\textsuperscript{*}  ~~ Jianshu Chen\textsuperscript{2} 
 ~~ \textbf{Dong Yu}\textsuperscript{2} \\
 \textsuperscript{1}Toyota Technological Institute at Chicago, Chicago, IL, USA \\
 \textsuperscript{2}Tencent AI Lab, Bellevue, WA, USA  \textsuperscript{3}Cornell, Ithaca, NY, USA \\
  haiwang@ttic.edu, ks985@cornell.edu, \\
  \{yudian,jianshuchen,dyu\}@tencent.com \\
}

\date{}

\begin{document}
	
\begin{CJK*}{UTF8}{gkai}

\maketitle

\blfootnote{* This work was done when H. W. and K. S. were at Tencent AI Lab, Bellevue, WA.}

\begin{abstract}

Recently, pre-trained language models have achieved remarkable success in a broad range of natural language processing tasks. However, in multilingual setting, it is extremely resource-consuming to pre-train a deep language model over large-scale corpora for each language. Instead of exhaustively pre-training monolingual language models independently, an alternative solution is to pre-train a powerful multilingual deep language model over large-scale corpora in hundreds of languages. However, the vocabulary size for each language in such a model is relatively small, especially for low-resource languages. This limitation inevitably hinders the performance of these multilingual models on tasks such as sequence labeling, wherein in-depth token-level or sentence-level understanding is essential.

In this paper, inspired by previous methods designed for monolingual settings, we investigate two approaches (\ie, joint mapping and mixture mapping) based on a pre-trained multilingual model BERT for addressing the out-of-vocabulary (OOV) problem on a variety of tasks, including part-of-speech tagging, named entity recognition, machine translation quality estimation, and machine reading comprehension. Experimental results show that using mixture mapping is more promising. To the best of our knowledge, this is the first work that attempts to address and discuss the OOV issue in multilingual settings. 

\end{abstract}
\section{Introduction}

% 1) oov problem; 
% 2) how people deal with it in monolingual settings
It has been shown that performance on many natural language processing tasks drops dramatically on held-out data when a significant percentage of words do not appear in the training data, \ie, out-of-vocabulary (OOV) words~\cite{sogaard2012robust,madhyastha2016mapping}. A higher OOV rate (\ie, the percentage of the unseen words in the held-out data) may lead to a more severe performance drop~\cite{kaljahi2015foreebank}.
OOV problems have been addressed in previous works under monolingual settings, through replacing OOV words with their semantically similar in-vocabulary words~\cite{madhyastha2016mapping, kolachina2017replacing} or using character/word information~\cite{kim2016character,kim2018learning,chen2018combining} or subword information like byte pair encoding (BPE)~\cite{sennrich2016neural,stratos2017sub}. %,li2018subword

% 3) bert (mono->multi), still oov limitations
Recently, fine-tuning a pre-trained deep language model, such as Generative Pre-Training (GPT)~\cite{radford2018improving} and Bidirectional Encoder Representations from
Transformers (BERT)~\cite{devlin2018bert}, has achieved remarkable success on various downstream natural language processing tasks. Instead of pre-training many monolingual models like the existing English GPT, English BERT, and Chinese BERT, a more natural choice is to develop a powerful multilingual model such as the multilingual BERT.

However, all those pre-trained models rely on language modeling, where a common trick is to tie the weights of softmax and word embeddings~\cite{press2017using}. Due to the expensive computation of softmax~\cite{yang2017breaking} and data imbalance across different languages, the vocabulary size for each language in a multilingual model is relatively small compared to the monolingual BERT/GPT models, especially for low-resource languages. Even for a high-resource language like Chinese, its vocabulary size $10$k in the multilingual BERT is only half the size of that in the Chinese BERT. Just as in monolingual settings, the OOV problem also hinders the performance of a multilingual model on tasks that are sensitive to token-level or sentence-level information. For example, in the POS tagging problem (Table \ref{tab:pos_result}), 11 out of 16 languages have significant OOV issues (OOV rate $\ge 5\%$) when using multilingual BERT.

%(1.1k per language on average)

% 4) motivation of solution
% (or re-train it from a pre-trained checkpoint),
According to previous work~\cite{radford2018improving,devlin2018bert}, it is time-consuming and resource-intensive to pre-train a deep language model over large-scale corpora. To address the OOV problems, instead of pre-training a deep model with a large vocabulary, we aim at enlarging the vocabulary size when we fine-tune a pre-trained multilingual model on downstream tasks. 
% Even with $110$k vocabulary size, it may take one year to pre-train a monolingual BERT with eight P100 GPUs~\cite{devlin2018bert}. 

% 5) contributions
We summarize our contributions as follows: (i) We investigate and compare two methods to alleviate the OOV issue. To the best of our knowledge, this is the first attempt to address the OOV problem in multilingual settings. (ii) By using English as an interlingua, we show that bilingual information helps alleviate the OOV issue, especially for low-resource languages. (iii) We conduct extensive experiments on a variety of token-level and sentence-level downstream tasks to examine the strengths and weaknesses of these methods, which may provide key insights into future directions\footnote{Improved models will be available at \url{https://github.com/sohuren/multilingul-bert}.}.

\section{Approach}
\label{sec:method}

We use the multilingual BERT as the pre-trained model. We first introduce the pre-training procedure of this model (Section~\ref{sec:approach:bert}) and then introduce two methods we investigate to alleviate the OOV issue by expanding the vocabulary (Section~\ref{sec:approach:vocab}). Note that these approaches are not restricted to BERT but also applicable to other similar models.

\subsection{Pre-Trained BERT}
\label{sec:approach:bert}

Compared to GPT~\cite{radford2018improving} and ELMo~\cite{peters2018deep}, BERT~\cite{devlin2018bert} uses a \emph{\bf bidirectional} transformer, whereas GPT pre-trains a left-to-right transformer~\cite{liu2018generating}; ELMo~\cite{peters2018deep} independently trains left-to-right and right-to-left LSTMs~\cite{peters2017semi} to generate representations as additional features for end tasks. 

In the pre-training stage,~\newcite{devlin2018bert} use two objectives: masked language model (LM) and next sentence prediction (NSP). In masked LM, they randomly mask some input tokens and then predict these masked tokens. Compared to unidirectional LM, masked LM enables representations to fuse the context from both directions. In the NSP task, given a certain sentence, it aims to predict the next sentence. The purpose of adding the NSP objective is that many downstream tasks such as question answering and language inference require sentence-level understanding, which is not directly captured by LM objectives.

\begin{figure*}[!ht]
\begin{center}
\includegraphics[width=0.94\textwidth]{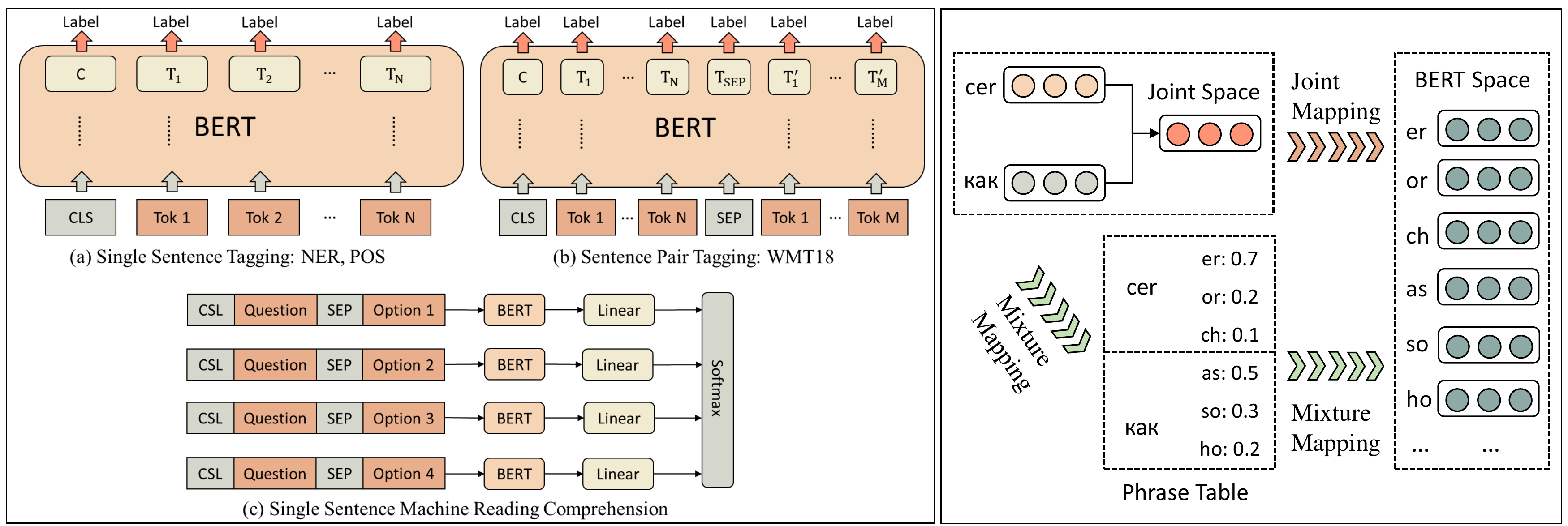}
\caption{Left: fine-tuning BERT on different kinds of end tasks. Right: illustration of joint and mixture mapping (in this example, during mixture mapping, we represent $\bm{e}(cer)=0.7*\bm{e}(er) + 0.2*\bm{e}(or) + 0.1*\bm{e}(ch)$).}
\label{fig:method:bertmerge}
\end{center}
\end{figure*}

After pre-training on large-scale corpora like Wikipedia and BookCorpus~\cite{zhu2015aligning}, we follow recent work~\cite{radford2018improving,devlin2018bert} to fine-tune the pre-trained model on different downstream tasks with minimal architecture adaptation. We show how to adapt BERT to different downstream tasks in Figure~\ref{fig:method:bertmerge} (left). 

%\todo{check whether the current figure 1 (c1, c2) is correct or not.}
%\todo{what we use for Chinese MRC is c1.}

\subsection{Vocabulary Expansion}
\label{sec:approach:vocab}

%https://www.aclweb.org/anthology/Q/Q17/Q17-1024.pdf
%https://arxiv.org/pdf/1703.03906.pdf
%https://arxiv.org/pdf/1609.08144.pdf%20(7.pdf

\newcite{devlin2018bert} pre-train the multilingual BERT on Wikipedia in $102$ languages, with a shared vocabulary that contains $110$k subwords calculated from the WordPiece model~\cite{wu2016google}. If we ignore the shared subwords between languages, on average, each language has a $1.1$k vocabulary, which is significantly smaller than that of a monolingual pre-trained model such as GPT ($40$k). The OOV problem tends to be less serious for languages (\eg, French and Spanish) that belong to the same language family of English. However, this is not always true, especially for morphologically rich languages such as German~\cite{ataman2018compositional,lample2018phrase}. OOV problem is much more severe in low-resource scenarios, especially when a language (\eg, Japanese and Urdu) uses an entirely different character set from high-resource languages.
%that has the largest Wikipedia since they share a large percentage of sub-words. 
% and English BERT ($30$k).

We focus on addressing the OOV issue at subword level in multilingual settings. Formally, suppose we have an embedding $E_{bert}$ extracted from the (non-contextualized) embedding layer in the multilingual BERT (i.e., the first layer of BERT). And suppose we have another set of (non-contextualized) sub-word embeddings $\{E_{l_{1}}, E_{l_{2}}, \dots, E_{l_{n}}\} \cup \{E_{en}\}$, which are pre-trained on large corpora using any standard word embedding toolkit. Specifically, $E_{en}$ represents the pre-trained embedding for English, and $E_{l_{i}}$ represents the pre-trained embedding for non-English language $l_{i}$ at the subword level. We denote the vocabulary of $E_{bert}$, $E_{en}$, and $E_{l_i}$ by $V_{bert}$, $V_{en}$, and $V_{l_{i}}$, respectively. For each subword $w$ in $V_{bert}$, we use $E_{bert}(w)$ to denote the pre-trained embedding of word $w$ in $E_{bert}$. $E_{l_i}(\cdot)$ and $E_{en}(\cdot)$ are defined in a similar way as $E_{bert}(\cdot)$. For each non-English language $l \in \{l_{1}, l_{2}, \dots, l_{n}\}$, we aim to enrich $E_{bert}$ with more subwords from the vocabulary in $E_{l_i}$ since $E_{l_i}$ contains a larger vocabulary of language $l_i$ compared to $E_{bert}$. 

%\todo{reviewer 3: motivation behind using these methods}
As there is no previous work to address multilingual OOV issues, inspired by previous solutions designed for monolingual settings, we investigate the following two methods, and all of them can be applied at both word/subword level, though we find subword-level works better (Section~\ref{sec:experiment}). %Though we find the later one works better in practice.

%which differs from \newcite{madhyastha2016mapping} in that they employ a non-linear mapping without imposing any constrains on the mapping, while 
%and $A_{l}^{T}=A_{l}^{-1}$\
% that makes it hard to incorporate the orthogonal constraints 反正这个也不work,可能不需要加这一句

\noindent\textbf{Joint Mapping} For each non-English language $l$, we first construct a joint embedding space $E'_{l}$ through mapping $E_{l}$ to $E_{en}$ by an orthogonal mapping matrix $B_{l}$ (\ie, $E'_{l}=E_{l}B_{l}$). When a bilingual dictionary $f_l:V_{l} \to V_{en}$ is available or can be constructed based on the shared common subwords (Section~\ref{subsec:expsetting}), we obtain $B_{l}$ by minimizing:
\begin{equation}
\small
\sum_{w'\in V_{l} \cap \{ w: f_{l}(w) \in V_{en}\}} \!\!\!\!\!\!\!\!\!\! \norm{E_{l}(w')B_{l} - E_{en}(f_l(w'))}_{F}^{2} \nonumber
\end{equation}
%%two languages share significant common words/subwords, 
%When a bilingual dictionary is not available and two languages does not share too much words/subword 
where $\norm{\cdot}_{F}$ denotes the Frobenius norm. Otherwise, for language pair (\eg, English-Urdu) that meets neither of the above two conditions, we obtain $B_{l}$ by an unsupervised word alignment method from MUSE~\cite{conneau2017word}.

% \footnote{We can also learn a single $A'$ jointly for all languages} removed??
We then map $E'_{l}$ to $E_{bert}$ by an orthogonal mapping matrix $A'_{l}$, which is obtained by minimizing
\begin{equation}
\small
\sum_{w\in f_{l}(V_{l}) \cap V_{bert}} \norm{E'_{l}(w)A'_{l} - E_{bert}(w)}_{F}^{2} \nonumber
\end{equation}
We denote this method by $M_J$ in our discussion below, where the subscript $J$ stands for ``joint''.
% This method, denoted by $M_{J}$, can be regarded as a joint version of $M_{I}$ since we first map all $E_{l}$ into a joint embedding space $E'_l$ defined by $E_{en}$.

\noindent\textbf{Mixture Mapping} Following the work of~\newcite{gu2018universal} where they use English as \emph{``universal tokens''} and map all other languages to English to obtain the subword embeddings, we represent each subword in $E'_{l}$ (described in joint mapping) as a mixture of English subwords where those English subwords are already in the BERT vocab $V_{bert}$. This method, denoted by $M_{M}$, is also a joint mapping without the need for learning the mapping from $E'_{l}$ to $E_{bert}$. Specifically, for each $w\in V_{l}$, we obtain its embedding $\bm{e}(w)$ in the BERT embedding space $E_{bert}$ as follows. 
\begin{equation}
\small
 \bm{e}(w) = \sum_{u \in \mathcal{T}(w)} p(u\,|\,w)E_{bert}(u) \nonumber %\label{eqn:mix1}
\end{equation}
where $\mathcal{T}(w)$ is a set to be defined later, and the mixture coefficient $p(u|w)$ is defined by
\begin{equation}
\small
    p(u\,|\,w) = \frac{\text{exp}(\text{CSLS}(E_{l}(u), E_{en}(w)))} {\sum_{v \in \mathcal{T}(w)}  \text{exp}(\text{CSLS}(E_{l}(v), E_{en}(w)))} \nonumber
\end{equation}%
where $\text{CSLS}$ refers to the distance metric Cross-domain Similarity Local Scaling~\cite{conneau2017word}. We select five $v \in V_{en}\cup V_{bert}$ with the highest $\text{CSLS}(E_{l}(v), E_{en}(w))$ to form set $\mathcal{T}(w)$. In all our experiments, we set the number of nearest neighbors in $\text{CSLS}$ to $10$. We refer readers to~\newcite{conneau2017word} for details. Figure~\ref{fig:method:bertmerge} (right) illustrates the joint and mixture mapping.

\section{Experiment}
\label{sec:experiment}

\subsection{Experiment Settings}
\label{subsec:expsetting}

We obtain the pre-trained embeddings of a specific language by training fastText~\cite{bojanowski2017enriching} on Wikipedia articles in that language, with context window $5$ and negative sampling $5$. Before training, we first apply BPE~\cite{sennrich2016neural} to tokenize the corpus with subword vocabulary size $50$k. For joint mapping method $M_J$, we use bilingual dictionaries provided by \newcite{conneau2017word}. For a language pair where a bilingual dictionary is not easily available, if two languages share a significant number of common subwords (this often happens when two languages belong to the same language family), we construct a bilingual dictionary based on the assumption that identical subwords have the same meaning~\cite{sogaard2018limitations}. We add all unseen subwords from $50$k vocabulary to BERT. We define a word as an OOV word once it cannot be represented as a single word. For example, in BERT, the sentence \emph{``Je sens qu' entre ça et les films de médecins et scientifiques"} is represented as \emph{``je sens qu \#\#' entre [UNK] et les films de [UNK] et scientifiques"}, where \textbf{qu'} is an OOV word since it can only be represented by two subword units: \textbf{qu} and \textbf{\#\#'}, but it is not OOV at subword level; \textbf{ça} and \textbf{médecins} cannot be represented by any single word or combination of subword units, and thus they are OOV at both word and subword level.    

% By default, 
%The first three tasks focus on token-level classification, and the last two focus on sentence-level classification.
We use the \textbf{multilingual BERT} with default parameters in all our experiments, except that we tune the batch size and training epochs. To have a thorough examination about the pros and cons of the explored methods, we conduct experiments on a variety of token-level and sentence-level classification tasks: part of speech (POS) tagging, named entity recognition (NER), machine translation quality estimation, and machine reading comprehension. See more details in each subsection.    % natural language inference (NLI),

\subsection{Discussions about Mapping Methods}
Previous work typically assumes a linear mapping exists between embedding spaces of different languages if their embeddings are trained using similar techniques~\cite{xing2015normalized,madhyastha2016mapping}. However, it is difficult to map embeddings learned with different methods~\cite{sogaard2018limitations}. Considering the differences between BERT and fastText: \eg, the objectives, the way to differentiate between different subwords, and the much deeper architecture of BERT, it is very unlikely that the (non-contextualized) BERT embedding and fastText embedding reside in the same geometric space. Besides, due to that BERT has a relatively smaller vocabulary for each language, when we map a pre-trained vector to its portion in BERT indirectly as previous methods, the supervision signal is relatively weak, especially for low-resource languages. In our experiment, we find that the accuracy of the mapping from our pre-trained English embedding to multilingual BERT embedding (English portion) is lower than $30\%$. In contrast, the accuracy of the mapping between two regular English embeddings that are pre-trained using similar methods (e.g., CBOW or SkipGram~\cite{mikolov2013distributed}) could be above $95\%$~\cite{conneau2017word}.

Besides the \textbf{joint mapping} method $M_J$ (Section~\ref{sec:approach:vocab}), another possible method that could be used for OOV problem in multilingual setting is that, for each language $l$, we map its pre-trained embedding space $E_{l}$ to embedding $E_{bert}$ by an orthogonal mapping matrix $A_{l}$, which is obtained by minimizing $\sum_{w\in V_{l} \cap V_{bert}} \norm{E_{l}(w)A_{l} - E_{bert}(w)}_{F}^{2}$. This approach is similar to~\cite{madhyastha2016mapping}, and is referred as \textbf{independent mapping} method below. However, we use examples to demonstrate why these kind of methods are less promising. In Table~\ref{tab:result:mapping}, the first two rows are results obtained by mapping our pre-trained English embedding (using fastText) to the (non-contextualized) BERT embedding. In this new unified space, we align words with CSLS metric, and for each subword that appears in English Wikipedia, we seek its closest neighbor in the BERT vocabulary. Ideally, each word should find itself if it exists in the BERT vocabulary. However, this is not always true. For example, although \emph{``however"} exists in the BERT vocabulary, independent mapping fails to find it as its own closest neighbor. Instead, it incorrectly maps it to irrelevant Chinese words ``盘'' (\emph{``plate''}) and ``龙'' (\emph{``dragon''}). A similar phenomenon is observed for Chinese. For example, ``册'' is incorrectly aligned to Chinese words ``书'' and ``卷''.

\begin{table}[!ht]
\centering
\footnotesize
\begin{tabular}{llll}
\toprule
Source Lang & Source & Target & probability \\ 
\midrule
\multirow{2}{*}{English}& however & 盘 (plate)  & 0.91\\
&however & 龙 (dragon) & 0.05\\
\midrule
\multirow{2}{*}{Chinese}& 册 (booklet) & 书 (book) & 0.49 \\
&册 (booklet) & 卷 (volume) & 0.46 \\
\bottomrule
\end{tabular}
\caption{Alignment from Independent Mapping.}
\label{tab:result:mapping}
\end{table}

Furthermore, our POS tagging experiments (Section~\ref{sec:mono_seq}) further show that joint mapping $M_{J}$ does not improve (or even hurt) the performance of the multilingual BERT. Therefore, we use \textbf{mixture mapping} $M_{M}$ to address and discuss OOV issues in the remaining sections.

%\todo{Reviewers may still complain about the early rejection of the other two mapping methods by just the POS tagging experiments. The reason is that we spent a lot of effort explaining all three methods, and this will give the reviewer the impression that we are going to do a thorough experiment study and comparision between all these three methods. Reviewer 1 is complaining about the lack of novelty in the modeling part. So we may pose our paper as an extensive experiment study and comparision of different candidate. Otherwise, if we reject the method too early and only use one method, then the reviewer will have the impression that we are just using one existing method to solve the problem, which seems to have less novelty in modeling. If time permits, I think it might be better to add the results for the other two mapping methods.}

% POS task
\begin{table*}[!ht]
\centering
\footnotesize
%\small
  \begin{tabular}{l|ccc|cccc|cc}
    \toprule
    %\textbf{Language}
    &
    \multicolumn{1}{c}{ $\text{BTS}^{\clubsuit}$ }  & \multicolumn{1}{c}{ $\text{BiLSTM}^{\diamondsuit}$ } & 
    \multicolumn{1}{c}{ $\text{FREQ}^{\diamondsuit}$ } &
    \multicolumn{1}{c}{\text{BERT}} & \multicolumn{1}{c}{$\text{BERT}_{\text{oov}}$} & \multicolumn{1}{c}{$\text{BERT}_{\text{oovR}}$} & 
    \multicolumn{1}{c}{$\text{BERT}_{\text{oovMJ}}$} &
    \multicolumn{1}{c}{ $\text{OOV}_{\text{w}}$} & \multicolumn{1}{c}{$\text{OOV}_{\text{sw}}$} \\
     % \midrule
     % Ave & 95.70 & 96.50 & 96.50 & & & \\
     %\midrule     
     %Indoeur &- & 96.63&96.61 & & & \\
     %non-Indo. &- &96.21 &96.28 & & & \\
     %Germanic &- &95.55 &95.49 & & & \\
     %Romance &- &96.93 &96.93 & & & \\
     %Slavic &- &97.42 &97.43 & & & \\
     \midrule
     ar &- & {98.23} & 90.06& 53.34 & \textbf{56.70} & 56.57  & 56.23 & 89.8 & 70.6 \\
     bg &97.84& 98.23& 90.06 & {\textbf{98.70}} & 98.22 & 94.41 & 97.21& 45.7 & 1.2 \\
     %cs &98.50 &98.02 &97.89 & & &  \\ 
     da &95.52 & 96.16& 96.35 & \textbf{97.16} & 96.53 & 94.15 &  94.85 & 38.9 & 2.8 \\
     de &92.87 & 93.51& 93.38 & 93.58 & \textbf{93.81} & 91.77  & 93.12 & 43.2 & 5.6 \\
     %en &93.87 & 95.17& 95.16 & 97.19 & - & - \\
     es &95.80 & 95.67& 95.74 & 96.04 & \textbf{96.92} & 95.10  & 95.77& 29.4 & 6.0\\
     %eu &- & 95.38 & 95.51 & 96.27 &  &  \\
     fa &96.82 & {97.60} & 97.49  & 95.62 & 94.90 & 94.35  & \textbf{95.82}& 35.6 & 6.5 \\
     fi &95.48 &95.74& {95.85} & 87.72 & \textbf{93.35} & 84.75  & 89.39& 64.9 & 10.4 \\
     fr &95.75 & 96.20 &96.11 & 95.17 & \textbf{96.59} & 94.84  & 95.24& 33.9 & 10.3 \\
     %he &- & 96.92 &96.96 & & &  \\
     %hi &- & 96.97 &97.10 & 80.70 & &  &\\
     hr &- & 96.27 & {96.82} & 95.03 & \textbf{96.49} & 89.87  & 93.48 & 49.5 & 8.3 \\
    % id &92.85 &93.32 &93.41 & 92.38 & 91.75 &  \\
     it &97.56 &97.90 &97.95 & \textbf{98.22} & 98.00 & 97.63  & 97.85& 30.3 & 2.3 \\
     nl &- &92.82 &93.30 & \textbf{93.89} & 92.89 & 91.30  & 91.71& 35.5 & 0.3\\
     no &- & {98.06} &98.03 & \textbf{97.25} & 95.98 & 94.21  & 95.83 & 38.7 & 4.4 \\
     pl &- & {97.63} &97.62 & 91.62 & \textbf{95.95} & 87.50 & 92.56 & 56.5 & 13.6 \\
     pt &- & {97.94} &97.90 & 96.66 & \textbf{97.63} & 95.93 & 96.90 & 34.0 & 8.3 \\
     sl &- & {96.97} &96.84 & 95.02 & \textbf{96.91} & 89.55 & 93.97& 49.2 & 7.8 \\
     sv &95.57 &96.60 & {96.69} & 91.23 & \textbf{96.66} & 90.45 & 91.92 & 48.2 & 17.7 \\
     \midrule
     average & - & 96.60 & 95.64 & 92.27 & 93.60 & 90.15 & 92.20 & 45.2 & 11.0 \\
\bottomrule
  \end{tabular}
  \caption{POS tagging accuracy (\%) on the Universal Dependencies v1.2 dataset. $\text{BERT}_{\text{oov}}$: BERT with method $M_{M}$. $\text{BERT}_{\text{oovR}}$: BERT with randomly picked embedding from BERT. $\text{BERT}_{\text{oovMJ}}$: BERT with method $M_{J}$. $\text{OOV}_{\text{w}}$: word-level OOV rate. $\text{OOV}_{\text{sw}}$: subword-level OOV rate. $\clubsuit$: \newcite{gillick2016multilingual}, $\diamondsuit$: \newcite{plank2016multilingual}.}
  \label{tab:pos_result}
 \end{table*}

\begin{table*}[!ht]
\centering
\footnotesize
  %\begin{tabular}{l ccccc|ccc}
  \begin{tabular}{l  ccc}
    \toprule
      %& \multicolumn{3}{c}{\textbf{Weibo NER}} \\
    \textbf{Approach}  & \bf Precision & \bf Recall & \bf F1 score \\
\midrule
DomainMask~\cite{peng2017multi}  & 60.8 & 44.9 & 51.7 \\
Linear Projection~\cite{peng2017multi} & 67.2 & 41.2 & 51.1 \\
Updates~\cite{peng2017supplementary}  & - & - & 56.1 \\
Updates~\cite{peng2017supplementary}  & - & -  & 59.0 \\
%\newcite{chen2015long}  & - & - & - \\
\midrule
BERT & 56.6 & 61.7 & 59.0 \\
$\text{BERT}_{\text{oov}} $ & 60.2 & 62.8 & \textbf{61.4} \\
$\text{BERT}_{\text{zh}} $ & 63.4 & 70.8 & \textbf{66.9} \\ 
\bottomrule
\end{tabular}
\caption{Performance of various models on the test set of Weibo NER. $\text{BERT}_{\text{zh}}$: Chinese BERT pre-trained over Chinese Wikipedia. We use scripts \textit{conlleval} for evaluation on NER.}
\label{tab:weiboner_result}
\end{table*}

\subsection{Monolingual Sequence Labeling Tasks}
\label{sec:mono_seq}
\textbf{POS Tagging}: We use the Universal Dependencies
v1.2 data~\cite{mcdonald2013universal}. For languages with token segmentation ambiguity, we use the gold segmentation following~\newcite{plank2016multilingual}. We consider languages that have sufficient training data and filter out languages that have unsatisfying embedding alignments with English (accuracy is lower than $20.0\%$ measured by word alignment accuracy or $0.25$ by unsupervised metric in MUSE~\cite{conneau2017word}). Finally, we keep $16$ languages. We use the original multilingual BERT (without using CRF~\cite{Lafferty:2001:CRF:645530.655813} on top of it for sequence labeling) to tune hyperparameters on the dev set and use the fixed hyperparameters for the expanded multilingual model. We do not tune the parameters for each model separately. As shown in Table~\ref{tab:pos_result}, at both the word and subword level, the OOV rate in this dataset is quite high. Mixture mapping improves the accuracy on $10$ out of $16$ languages, leading to a $1.97\%$ absolute gain in average. We discuss the influence of alignments in Section~\ref{sec:discussion}.

%See Section~\ref{sec:discussion} for more details of the influence of alignments.

% we obtain a xx absolute improvement in average accuracy
%in average accuracy improvement 2.56 and average decrease XX. 

%If there is more than one treebank per language, we use the treebank that has the canonical language name (e.g., Finnish instead of Finnish-FTB). 
%check \cite{sagot2017improving}, they also improve a little bit, but on different datasets.
%\cite{plank2016multilingual}, 
%survey (suggested some additional datasets, but mostly for english): %\cite{changpinyo2018multi} has some results on Ja language etc \cite{kemos2018neural}
%used UD2.0 \cite{kann2018character} \cite{yu2017general} has some better results on some languages.

%\todo{report FOOV}
%\todo{add discussions about the random initialized OOV baseline?}
%\todo{double-check the state-of-the-art results for Chinese segmentation}

\noindent\textbf{Chinese NER}: We are also interested in investigating the performance gap between the expanded multilingual model and a monolingual BERT that is pre-trained on a large-scale monolingual corpus. Currently, pre-trained monolingual BERT models are available in English and Chinese. As English has been used as the interlingua, we compare the expanded multilingual BERT and the Chinese BERT on a Chinese NER task, evaluated on the Weibo NER dataset constructed from social media by~\newcite{peng2015named}. In the training set, the token-level OOV rate is $2.17\%$, and the subword-level OOV rate is $0.54\%$. We tune the hyperparameters of each model based on the development set separately and then use the best hyperparameters of each model for evaluation on the test set. 

As shown in Table~\ref{tab:weiboner_result}, the expanded model outperforms the multilingual BERT on the Weibo NER dataset. We boost the F1 score from $59.0\%$ to $61.4\%$. Compared to the Chinese BERT ($66.9\%$), there still exists a noticeable performance gap. One possible reason could be the grammatical differences between Chinese and English. As BERT uses the language model loss function for pre-training, the pre-trained Chinese BERT could better capture the language-specific information comapred to the multilingual BERT.

\subsection{Code-Mixed Sequence Labeling Tasks}
\label{sec:mixed}

As the multilingual BERT is pre-trained over $102$ languages, it should be able to handle code-mixed texts. Here we examine its performance and the effectiveness of the expanded model in mixed language scenarios, using two tasks as case studies. 

\noindent\textbf{Code-Switch Challenge}: We first evaluate on the CALCS-2018 code-switched task~\cite{calcs2018shtask}, which contains two NER tracks on Twitter social data: mixed English\&Spanish (en-es) and mixed Modern Standard Arabic\&Egyptian (ar-eg). Compared to traditional NER datasets constructed from news, the dataset contains a significant portion of uncommon tokens like hashtags and abbreviations, making it quite challenging. For example, in the en-es track, the token-level OOV rate is $44.6\%$, and the subword-level OOV rate is $3.1\%$; in the ar-eg track, the token-level OOV rate is $64.0\%$, and the subword-level OOV rate is $6.0\%$. As shown in Table~\ref{tab:mixed_result:twitter}, on ar-eg, we boost the F1 score from $74.7\%$ to $77.3\%$. However, we do not see similar gains on the en-es dataset, probably because that English and Spanish share a large number of subwords, and adding too many new subwords might prevent the model from utilizing the well pre-trained subwords embedding. See Section~\ref{sec:discussion} for more discussions.

%\footnote{Note that our evaluation is based on dev set since the test set is only open to participants}
%we boost the precision from $72.7$ to $74.2$ with recall drops, we think this is due to English and Spanish share subwords significantly .
%, adding too much subwords will let the model learn better embedding for long subwords (precision boost) but prevent the model learn embedding for short subword, and short subword is important for unseen words (recall drop). 

%on SIGHAN CWS dataset, we boost F1 score from 95.5 to 95.6. Of course, compared with BERT Chinese model, there is still some gap. We suspect this is due to Chinese has different language order compared with English, thus pure Chinese BERT model          

\eat{
\cite{indra2018bilingual} code switch on en and es, method is simple and potentially we can use the data.
\cite{aguilar2018named}
train: 
en-es:
OOV@token: 0.446437
OOV@subword: 0.030605
en-es (enriched)
OOV@token: 0.416830
OOV@subword: 0.013943
train:
ar-et:
OOV@token: 0.64
OOV@subword: 0.06
ar-et(enriched)
OOV@token:0.54 
OOV@subword:0.00888
}

\begin{table}[!ht]
\centering
%\small
\footnotesize
\begin{tabular}{lcccccccc}
\toprule
     & \multicolumn{3}{c}{\textbf{en-es}} & \multicolumn{3}{c}{\textbf{ar-eg}} \\
    \textbf{Model} & Prec & Rec & F1 & Prec & Rec & F1 \\ 
    \midrule
    %\multicolumn{2}{l}{Test} & \\ 
    $\text{FAIR}^{\clubsuit}$ & - & - & 62.4 & - & - & 71.6 \\ 
    $\text{IIT}^{\clubsuit}$ & - & - & 63.8 & -& - & - \\
    \midrule
    %\multicolumn{2}{l}{Dev} & \\
    $\text{FAIR}^{\diamondsuit}$ & - & - & 67.7 & - & - & \textbf{81.4} \\
    BERT & 72.7 & \textbf{63.6} &\textbf{67.8} & 73.8 & 75.6 & 74.7 \\ 
    $\text{BERT}_{\text{oov}}$ & \textbf{74.2} & 60.9 & 66.9 & \textbf{76.9} & \textbf{77.8} & \textbf{77.3}\\
    \bottomrule
  \end{tabular}
  \caption{Accuracy (\%) on the code-switch challenge. The top two rows are based on the test set, and the bottom three rows are based on the development set. $\clubsuit$: results from~\newcite{calcs2018shtask}. $\diamondsuit$: results from~\newcite{wang2018code}. }
  \label{tab:mixed_result:twitter}
 \end{table}

\noindent \textbf{Machine Translation Quality Estimation:}
 All previous experiments are based on well-curated data. Here we evaluate the expanded model on a language generation task, where sometimes the generated sentences are out-of-control. 

 We choose the automatic Machine Translation Quality Estimation task and use Task $2$ -- word-level quality estimation -- in WMT18~\cite{wmt2018}. Given a source sentence and its translation (\ie, target), this task aims to estimate the translation quality (\emph{``BAD"} or \emph{``OK"}) at each position: \eg, each token in the source and target sentence, each \textbf{gap} in the target sentence. We use English to German (en-de) SMT translation. On all three categories, the expanded model consistently outperforms the original multilingual BERT (Table~\ref{tab:mixed_result:wmt})\footnote{Our evaluation is based on the development set since the test set is only available to participants, and we could not find the submission teams' performance on the developmenet set.}.
 
\begin{table*}[!ht]
\centering
%\small
\footnotesize
\begin{tabular}{l ccc ccc ccc}
\toprule
     & \multicolumn{3}{c}{\textbf{Words in MT}} & \multicolumn{3}{c}{\textbf{Gaps in MT}} &\multicolumn{3}{c}{\textbf{Words in SRC}} \\
    \textbf{Model} & F1-BAD & F1-OK & F1-multi & F1-BAD & F1-OK & F1-multi & F1-BAD & F1-OK & F1-multi \\ 
    \midrule
    %\multicolumn{3}{l}{\textbf{En $\to$ De}} &  \\
    \newcite{fan2018bilingual} & 0.68 & 0.92 & \textbf{0.62} &- & - & -& -& - & -   \\ 
    \newcite{fan2018bilingual} & 0.66 & 0.92 & 0.61 & 0.51 & 0.98 & \textbf{0.50} &-  &- &- \\ 
    $\text{SHEF-PT}^{\clubsuit}$ & 0.51 & 0.85 & 0.43 & 0.29 & 0.96 & 0.28 & 0.42 & 0.80 & 0.34 \\
    \midrule
    BERT & 0.58 & 0.91 & 0.53 & 0.47 & 0.98 & 0.46 & 0.48 & 0.90 & 0.43   \\ 
    $\text{BERT}_{\text{oov}}$ & 0.60 & 0.91 & \textbf{0.55} & 0.50 & 0.98 & \textbf{0.49} & 0.49 & 0.90 & \textbf{0.44}   \\ 
    %\midrule
    %\multicolumn{3}{l}{\textbf{De $\to$ En}} &  \\
    %\newcite{fan2018bilingual} & 0.65 & 0.92 & 0.60 & - & - & - & - & - & -   \\ 
    %\newcite{fan2018bilingual} & 0.65 & 0.92 & 0.59 & - & - & - & - & -& -\\ 
    %CMU-LTI~\cite{specia2018findings} & 0.49 & 0.90 & 0.44 & - & - & - & - & - & - \\
    %Baseline~\cite{specia2018findings} & 0.49 & 0.87 & 0.42 & 0.21 & 0.97 & 0.2 & 0.39 & 0.89 & 0.35 \\
    %BERT base(mu) &  & &  & &  & & &  &    \\ 
    %BERT base(mu_oov) &  & &  & &  & & &  &    \\ 
    \bottomrule
  \end{tabular}
  \caption{WMT18 Quality Estimation Task 2 for the en$\to$de SMT dataset. $\clubsuit$: result from \newcite{specia2018findings}. \textbf{MT}: machine translation, \eg, target sentence, \textbf{SRC}: source sentence. F1-OK: F1 score for \emph{``OK"} class; F1-BAD: F1 score for \emph{``BAD"} class; F1-multi: multiplication of F1-OK and F1-BAD.}
  \label{tab:mixed_result:wmt}
 \end{table*}

% not significant xnli
\subsection{Sequence Classification Tasks}
\label{sec:xnli}
 
Finally, we evaluate the expanded model on sequence classification in a mixed-code setting, where results are less sensitive to unseen words.  % monolingual/

\noindent\textbf{Code-Mixed Machine Reading Comprehension}:
We consider the mixed-language machine reading comprehension task. Since there is no such public available dataset, we construct a new Chinese-English code-mixed machine reading comprehension dataset based on 37,436 unduplicated utterances obtained from the transcriptions of a Chinese and English mixed speech recognition corpus King-ASR-065-1\footnote{http://kingline.speechocean.com.}. We generate a multiple-choice machine reading comprehension problem (\ie, a question and four answer options) for each utterance. A question is an utterance with an English text span removed (we randomly pick one if there are multiple English spans) and the correct answer option is the removed English span. Distractors (\ie, wrong answer options) come from the top three closest English text spans, which appear in the corpus, based on the cosine similarity of word embeddings trained on the same corpus. For example, given a question ``突然听到~21\underline{~~~~~}，那强劲的鼓点，那一张张脸。'' (\emph{``Suddenly I heard 21\underline{~~~~~}, and the powerful drum beats reminded me of the players.''}) and four answer options \{ \emph{``forever''}, \emph{``guns''}, \emph{``jay''}, \emph{``twins'' }\}, the task is to select the correct answer option \emph{``guns''} (\emph{``21 Guns''} is a song by the American rock band \emph{Green Day}). We split the dataset into training, development, and testing of size 36,636, 400, 400, respectively. Annotators manually clean and improve the quality problems by generating more confusing distractors in the dev and testing sets to guarantee that these problems are error-free and challenging.
%\footnote{Annotations are available at \url{https://github.com/nlpdata/code-mixed}.}. 
% \footnote{We will release the code/annotations upon publication.}

In this experiment, for each BERT model, we follow its default hyperparameters. As shown in Table~\ref{tab:mixed_result:RC}, the expanded model improves the multilingual BERT ($38.1\%$) by $1.2\%$ in accuracy. Human performance ($81.4\%$) indicates that this is not an easy task even for human readers.

% in average
% gensim
% ceiling: (300 * 0.8875 + 500 * 0.9125) / 800 = 0.903125
% Human: (300 * 0.7125 + 500 * 0.875) / 800 = 0.8140625

\begin{table}[!ht]
\centering
%\small
\footnotesize
\begin{tabular}{lcc}
\toprule
     & \multicolumn{2}{c}{\textbf{Accuracy}} \\
    \textbf{Model} & Development & Test \\ 
     \midrule
    $\text{BERT}_{\text{en}}$ & 38.2 & 37.3   \\ 
    BERT & 38.7 & 38.1   \\ 
    \midrule
    $\text{BERT}_{\text{oov}}$ & 39.4 & \textbf{39.3} \\
    $\text{BERT}_{\text{zh}}$ & 40.0 & \textbf{45.0} \\
    %\midrule
    %Human Performance   &  &  \\
    %Ceiling Performance &  &  \\ 
    \bottomrule
  \end{tabular}
  \caption{Accuracy (\%) of models on the code-mixed reading comprehension dataset. $\text{BERT}_{\text{en}}$: pre-trained English BERT. $\text{BERT}_{\text{zh}}$: pre-trained Chinese BERT.}
  \label{tab:mixed_result:RC}
 \end{table}

%---------------------------------------------D---------------------------------------------------%

\subsection{Discussions}
\label{sec:discussion}

In this section, we first briefly investigate whether the performance boost indeed comes from the reduction of OOV and then discuss the strengths and weaknesses of the methods we investigate. 

First, we argue that it is essential to alleviate the OOV issue in multilingual settings. Taking the POS tagging task as an example, we find that most errors occur at the OOV positions (Table~\ref{tab:pos_result:analysis} in Section~\ref{sec:mono_seq}). In the original BERT, the accuracy of OOV words is much lower than that of non-OOV words, and we significantly boost the accuracy of OOV words with the expanded BERT. All these results indicate that the overall improvement mostly comes from the reduction of OOV.

\begin{table}[!ht]
\centering
%\small
\footnotesize
\begin{tabular}{lcccc}
\toprule
    & \multicolumn{2}{c}{BERT} & \multicolumn{2}{c}{$\text{BERT}_{\text{oov}}$} \\
    \textbf{Lang} & non-OOV & OOV & non-OOV & OOV \\ 
    \midrule
     %ar &  &  &  &    \\ 
     %de &  &  &  &     \\ 
     %es &  &  &  &    \\
     fi & 98.1 & 81.3 & 98.5 & 90.2   \\
     fr & 97.0 & 90.2 & 97.2 & 95.6   \\
     hr & 97.8 & 91.9 & 97.7 & 94.5   \\
     pl & 98.8 & 84.6 & 99.0 & 93.2   \\
     pt & 98.8 & 91.5 & 98.6 & 94.8   \\
     sl & 98.6 & 91.6 & 98.7 & 95.1   \\
     sv & 97.4 & 82.9 & 98.2 & 94.8    \\
     \midrule
     average & 98.1 & 87.7 & 98.3 & 94.0    \\
  \bottomrule
  \end{tabular}
  \caption{POS tagging accuracy (\%) for OOV tokens and non-OOV tokens on the Universal Dependencies v1.2 dataset, where the OOV/non-OOV are defined at word level with the original BERT vocabulary.}
  \label{tab:pos_result:analysis}
 \end{table}
 
%In our method, we mainly need to consider the following two sub-problems: what subwords to be added; how to obtain good representation for the newly added subwords. And we found the follow factors can influence those sub-problems.

We also observe that the following factors may influence the performance of the expanded model.

%the performance of our expanded model can be influenced by the following factors:
%the following factors can influence the performance of our expanded model.

\noindent\textbf{Subwords}: When expanding the vocabulary, it is critical to add only frequent subwords. Currently, we add all unseen subwords from the $50$k vocabulary (Section \ref{subsec:expsetting}), which may be not an optimal choice. Adding too many subwords may prevent the model from utilizing the information from pre-trained subword embedding in BERT, especially when there is a low word-level overlap between the training and test set. 

\noindent\textbf{Language}: We also find that languages can influence the performance of the vocabulary expansion through the following two aspects: the alignment accuracy and the closeness between a language and English. For languages that are closely related to English such as French and Dutch, it is relatively easy to align their embeddings to English as most subword units are shared~\cite{sogaard2018limitations,conneau2017word}. In such case, the BERT embedding already contains sufficient information, and therefore adding additional subwords may hurt the performance. On the other hand, for a distant language such as Polish (Slavic family), which shares some subwords with English (Germanic family), adding subwords to BERT brings performance improvements. In the meantime, as Slavic and Germanic are two subdivisions of the Indo-European languages, we find that the embedding alignment methods perform reasonably well. For these languages, vocabulary expansion is usually more effective, indicated by POS tagging accuracies for Polish, Portuguese, and Slovenian (Table~\ref{tab:pos_result}). For more distant languages like Arabic (Semitic family) that use different character sets, it is necessary to add additional subwords. However, as the grammar of such a language is very different from that of English, how to accurately align their embeddings is the main bottleneck. 

\noindent\textbf{Task}: We see more significant performance gains on NER, POS and MT Quality Estimation, possibly because token-level understanding is more critical for these tasks, therefore alleviating OOV helps more. In comparison, for sequence level classification tasks such as machine reading comprehension (Section~\ref{sec:xnli}), OOV issue is less severe since the result is based on the entire sentence.

\section{Related Work}

\hhide{
\subsection{Generative Pre-Training}
Great success has been achieved by pre-training deep neural networks with language modeling and then fine-tuning the models on downstream tasks~\cite{radford2018improving,devlin2018bert}. It significantly outperforms methods that use pre-trained language models as auxiliary features~\cite{peters2018deep,chen2018convolutional}. 
%Instead of using LSTM models~\cite{dai2015semi,peters2018deep,howard2018universal}, transformer networks~\cite{vaswani2017attention,liu2018generating} have become a popular choice to capture long-range structures.

However, it's time-consuming and resource-intensive to pre-train a deep language model such as GPT~\cite{radford2018improving} and BERT~\cite{devlin2018bert}. Considering the computation limitation, it is necessary to pre-train a multilingual model such as Multi-BERT (Section~\ref{sec:approach:bert}) that leverage rich linguistic knowledge from large-scale corpus in different languages. However, the vocabulary size is inevitably limited. To get a more powerful pre-trained multilingual model, instead of training from scratch, in this paper, we focus on improving it by enlarging the vocabulary size during the fine-tuning stage. 
}

%instead of using the vocabulary from the pre-training stage, which is inevitably limited by imbalanced data, computationally expensive softmax layers, and enormous computational resources.

%\subsection{Handling Out-Of-Vocabulary Words}
OOV poses challenges for many tasks~\cite{pinter2017mimicking} such as machine translation~\cite{razmara2013graph,sennrich2016neural} and sentiment analysis~\cite{kaewpitakkun2014sentiment}. Even for tasks such as machine reading comprehension that are less sensitive to the meanings of each word, OOV still hurts the performance~\cite{chu2017broad,zhang2018subword}. We now discuss previous methods in two settings.        

\subsection{Monolingual Setting}
%\noindent\textbf{Monolingual}:
Most previous work address the OOV problems in monolingual settings. Before more fine-grained encoding schema such as BPE~\cite{sennrich2016neural} is proposed, prior work mainly focused on OOV for token-level representations~\cite{taylor2011towards,kolachina2017replacing}. Besides simply assigning random embeddings to unseen words~\cite{dhingra2017gated} or using an unique symbol to replace all these words with a shared embedding~\cite{hermann2015teaching}, a thread of research focuses on refining the OOV representations based on word-level information, such as using similar in-vocabulary words~\cite{luong2015addressing,chousing2015,tafforeau2015adapting,li2016towards}, mapping initial embedding to task-specific embedding~\cite{rothe2016ultradense,madhyastha2016mapping}, using definitions of OOV words from auxiliary data~\cite{long2016leveraging,bahdanau2017learning}, and tracking contexts to build/update representations~\cite{henaff2016tracking,kobayashi2017neural,ji2017dynamic,zhao2018addressing}.

Meanwhile, there have been efforts in representing words by utilizing character-level~\cite{zhang2015character,ling2015finding,ling2015character,kim2016character,gimpelcharagram2016} or subword-level representations~\cite{sennrich2016neural,bojanowski2017enriching}. To leverage the advantages in character and (sub)word level representation, some previous work combine (sub)word- and character-level representations~\cite{santos2014learning,dos2015boosting,yu2017joint} or develop hybrid word/subword-character architectures~\cite{chung2016character,luong2016achieving,pinter2017mimicking,bahdanau2017learning,matthews2018using,li2018subword}. However, all those approaches assume monolingual setting, which is different from ours.
%Besides the fine-grained units, some work encode rare words with less interpretable symbols such as Huffman codes~\cite{chitnis2015variable} or unicode bytes~\cite{gillick2016multilingual}.

%\noindent\textbf{Multilingual}: 
\subsection{Multilingual Setting}
Addressing OOV problems in a multilingual setting is relatively under-explored, probably because most multilingual models use separate vocabularies~\cite{jaffe2017generating,platanios2018contextual}. While there is no direct precedent, previous work show that incorporating multilingual contexts can improve monolingual word embeddings~\cite{zou2013bilingual,andrew2013deep,faruqui2014improving,lu2015deep,ruder2017survey}.   

\newcite{madhyastha2017learning} increase the vocabulary size for statistical machine translation (SMT). Given an OOV source word, they generate a translation list in target language, and integrate this list into SMT system. Although they also generate translation list (similar with us), their approach is still in monolingual setting with SMT. \newcite{cotterell2017cross} train char-level taggers to predict morphological taggings for high/low resource languages jointly, alleviating OOV problems to some extent. In contrast, we focus on dealing with the OOV issue at subword level in the context of pre-trained BERT model.

\section{Conclusion}

We investigated two methods (\ie, joint mapping and mixture mapping) inspired by monolingual solutions to alleviate the OOV issue in multilingual settings. Experimental results on several benchmarks demonstrate the effectiveness of mixture mapping and the usefulness of bilingual information. To the best of our knowledge, this is the first work to address and discuss OOV issues at the subword level in multilingual settings. Future work includes: investigating other embedding alignment methods such as Gromov-Wasserstein alignment~\cite{alvarez2018gromov} upon more languages; investigating approaches to choose the subwords to be added dynamically.

%We find bilingual information is useful and mixture mapping outperforms other methods.
%, and we found bi-lingual information helps solving the OOV problem. 

%\newpage
%\clearpage

\section*{Acknowledgments}
We thank the anonymous reviewers for their encouraging and helpful feedback.

\bibliography{conll-2019}
\bibliographystyle{acl_natbib}

\end{CJK*}
\end{document}